# AN INVESTIGATION OF THE SEQUENTIAL SAMPLING METHOD FOR CROSSDOCKING SIMULATION OUTPUT VARIANCE REDUCTION

*Adrian Adewunmi*
*Prof. Uwe Aickelin*
School of Computer Science
University of Nottingham
Nottingham
aqa@cs.nott.ac.uk, uwe.aickelin@nottingham.ac.uk

*Dr. Mike Byrne*
Nottingham University Business School
University of Nottingham
Nottingham
Mike.Byrne@nottingham.ac.uk

**ABSTRACT:**
*This paper investigates the reduction of variance associated with a simulation output performance measure, using the Sequential Sampling method while applying minimum simulation replications, for a class of JIT (Just in Time) warehousing system called crossdocking. We initially used the Sequential Sampling method to attain a desired 95% confidence interval half width of ± 0.5 for our chosen performance measure (Total usage cost, given the mean maximum level of £157,000 and a mean minimum level of £149,000). From our results, we achieved a 95% confidence interval half width of ± 2.8 for our chosen performance measure (Total usage cost, with an average mean value of £115,000). However, the Sequential Sampling method requires a huge number of simulation replications to reduce variance for our simulation output value to the target level. Arena® (version 11) simulation software was used to conduct this study.*

## 1. INTRODUCTION

Many systems in areas such as manufacturing, warehousing and distribution can be too complex to model analytically (Buzacott and Yao, 1986). In particular, JIT (Just in Time) warehousing systems such as crossdocking can present such difficulty. This is because crossdocking distribution systems operate processes which exhibit an inherent random behaviour that can affect its overall expected performance. Crossdocking is a "process where product is received in a facility, occasionally married with other products going to the same destination, then shipped at the earliest opportunity, without going into long term storage" Napolitano (2000).

Generally, such crossdocking processes include fulfilment of incoming and outgoing orders by manual order picking operators and automated order picking machines. These order picking resources are usually available in shifts, constrained by capacity and scheduled into order picking jobs. There is also the potential that manual order picking operators have different skill levels and there is a possibility for automated order picking machines to breakdown. In such a situation, it becomes important for the achievement of a smooth crossdocking operation, to pay particular attention to the order picking process within the crossdocking distribution system Li et al (2004). The order picking process essentially needs to be fulfilled with minimal interruptions and with the least amount of resource cost (Lin and Lu, 1999).

A suitable technique for modelling and analysing complex systems like crossdocking is discrete event simulation Rohrer (1995). It has the ability to evaluate and measure the performance of a variety of output performance measure values of a crossdocking distribution centre Magableh et al (2005) and as a result, give us an insight into its random behaviour i.e. the sources of such model randomness. Using discrete event simulation involves utilising probabilistic distributions as part of the input parameter estimation; thus this will result in some variance associated with the output performance measure value. The greater the level of variance in the output value, the lower the precision the simulation results will contain Kelton et al (2007). Within our simulation model, there are different sources of variance; these include order arrival time and processing time which are based on probabilistic distributions.

One way to measure the performance of a crossdocking simulation model is by using the confidence interval half width on the selected performance measure Kelton et al (2007). We can identify high levels of variance associated with performance output value by a wider confidence interval half width Creda (1995). So for a better precision in simulation output value, and a higher confidence in the conclusions obtained, variance has to be reduced.

An accepted method to achieve variance reduction is to keep the simulation running until it reaches the desired confidence interval half width for the selected output performance measure. This can require a lot of computational effort to achieve a target solution quality Sasser et al (1970). We anticipate that by successively running the crossdocking simulation model, we can endeavour to answer two main questions:

- Is the variance within the model as a consequence of the characteristic of discrete event simulation input, i.e. the use of probability distributions and the use of random numbers or due to the random nature of the crossdocking distribution process or a combination of both?

- To what extent do the use of probability distribution on simulation model input and the use of random numbers influence the number of simulation replications required to achieve a target confidence interval half width for a selected output performance measure?

The diagram below illustrates a typical crossdocking distribution centre, highlighting the order picking area.

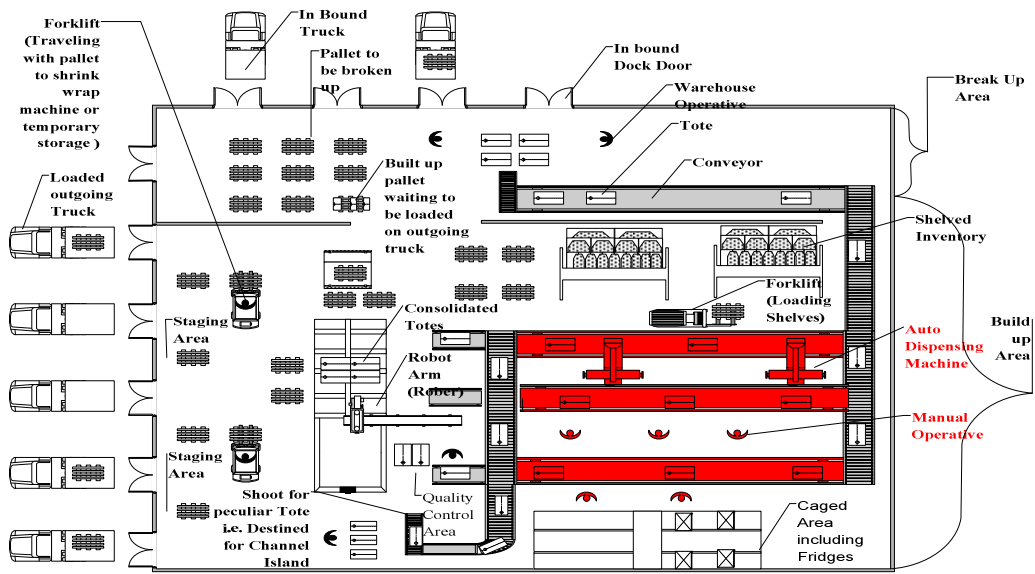

Figure 1. A typical crossdocking distribution centre.

## 2. BACKGROUND

For the design and analysis of complex models, where the performance of such models is measured using confidence interval half width on mean values, it is sometimes difficult to achieve a target precision at an acceptable computational cost because of variance. This variance can be due to the inherent randomness of the complex model under study or the technique applied in designing and analysing such models Wilson (1984). The kinds of models we are paying particular attention to are complex discrete event simulation models of dynamic systems i.e. the crossdocking distribution system.

Such models are often characterised by having one or more random input which are transformed by the complex method of discrete event simulation into one or more random output. Thus there is a need to apply appropriate statistical techniques to the random output for there to be a satisfactory level of confidence in the conclusions obtained through them. These statistical techniques are called variance reduction techniques Kelton et al (2007).

"The purpose of a variance reduction technique is to produce at no extra cost a more accurate estimate than that obtained without the application of the technique or to produce at less cost an estimate having the same accuracy. The means of the estimates obtained with and without application of the techniques should be equal and the accuracy of an estimate is measured by its variance" (Lavenberg and Welsh, 1978). An estimate being a simulation output performance measure that is of interest to the simulator.

There are a variety of techniques for reducing variance associated with an output performance measure, resulting from the evaluation of the performance of complex systems when using discrete event simulation McGeoch (1992). The main techniques are: Antithetic Variates and Control Variates Grant (1983), Importance Sampling and Stratified Sampling Glasserman et al (2000), Common Random Numbers Banks (1998) and the Sequential Sampling Method Kelton et al (2007).

The variance reduction techniques of interest are the Common Random Numbers and the Sequential Sampling method. Common Random Numbers entails dedicating a different stream of random numbers to each source of model randomness. The methodology under consideration here is the Sequential Sampling method. It is based on the principle of "sequentially determining the length of a single simulation run needed to construct an acceptable confidence interval for a steady state mean" (Law and Kelton ,2000). For a background treatment to the use of Common Random Number, and the Sequential Sampling Method as techniques for variance reduction, refer to Kelton et al (2007).

It is seldom possible to rely on an analytical procedure, to produce a fixed number of simulation replications, which will achieve a confidence interval half width on a selected output performance measure. This implies that before running the simulation model, one cannot be sure, how valid and precise the output will be or to estimate in advance the number of simulation replications necessary to yield the desired confidence interval half width. The dilemma is that an analyst has little idea in advance of the number of simulation replications that will yield a particular confidence interval half width on the selected output performance measure and as such will end up having to deal with the imprecision in the simulation output results. The Sequential Sampling method (Law and Carson, 1979, Kelton et al, 2007) causes the simulation model to keep replicating until it reaches the target confidence interval half width on the selected output performance measure.

We are interested in the Sequential Sampling method as a technique for variance reduction because after each replication, it checks to see if the desired confidence interval half width of a simulation output performance measure has been achieved, it also provides an opportunity to improve on existing techniques for reducing variance "on line" i.e. while the simulation replications are being performed." It is typically crucial to process data on-line as it arrives, both from the point of view of storage costs as well as for rapid adaptation to changing signal characteristics" Arulampalam et al (2002). "If we can somehow reduce the variance of an output random variable of interest, without disturbing its expectation, we can obtain better precision, e.g., smaller confidence intervals, for the same amount of simulation, or, alternatively achieve a desired precision with less simulating" (Law and Kelton ,2000).

Our experiments have been conducted using the Arena® Simulation Software, which calculates a 95% confidence interval half width for the expected output performance measure value across a specified number of replications over a single simulation run, using a method called the batch means. *"If there was enough data to perform the test for uncorrelated batches and the test were passed, the half width (the "plus-or-minus" amount) of a 95% confidence interval on the expected value of the statistic would be given"* Kelton et al (2007).

This paper continues a description of the research and study, and a conclusion which includes future work.

## 3. RESEARCH STUDY

### 3.1 RESEARCH PROBLEM

To reduce variance through the Sequential Sampling method using minimal simulation replications in order to achieve a target simulation output performance measure i.e. a 95% confidence interval half width of ± 0.5 on total usage cost given the mean maximum level of £157,000 and a mean minimum level of £149,000. Total usage cost measures the overall cost of both manual and automated resource usage associated with the order picking process for the crossdocking distribution centre.

### 3.2 METHODOLOGY

The Sequential Sampling method involves simulating one replication at a time until a predefined confidence interval half width is achieved Kelton et al (2007). Within the Arena® simulation software, the Sequential Sampling method parameters are decided before a simulation run. This is done by setting a large number of replications i.e. 999,999, then making a choice regarding the target half width and selecting the output performance measure of interest. After each replication, during the simulation run, the Sequential Sampling "logic" will check to see if the target half width has been achieved. If so, it will cause the simulation model to stop otherwise the simulation model will continue performing more replications.

Here is a brief summary of the flow of activities of the Sequential Sampling logic Kelton (2007):

- Entities are created and sent into a decide module,
- The decide module will check to see if the number of replications (NREP), is less than or equal to 2 (NREP is the number of initial simulation replications) and check that the target half width (ORUNHALF) has been achieved.
- If yes, the entity will be sent to the dispose module and the simulation will stop, otherwise it will keep replicating till the target half width has been achieved.

Below is the implementation of the Sequential Sampling method in the Arena® simulation software.

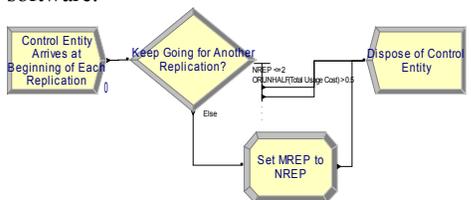

Figure 2. Arena® Terminating Sequential – Sampling Control Logic®

### 3.3 DATA COLLECTION

We collected sufficient procedural information and some basic data from a visit to a crossdocking distribution centre. This provided enough knowledge about the operation of an order picking process within a crossdocking distribution centre. This was very useful in determining the scope and level of detail required for the development of the simulation model.

Here is a description of the order picking process within a "typical" crossdocking distribution centre:

a. The order picking process is fulfilled by two types of resources, namely, the manual order picking operatives and the automated order dispensing machines.
b. A series of parallel conveyor belts run through the order picking area. These conveyor belts take "totes" (Plastic containers which hold picked order items) past the automated order dispensing machines and manual order picking operatives at a reasonable speed.
c. The manual order pickers walk short distances to pick and fill items into totes on the conveyor, while the automated order dispensing machine performs its role by delaying the totes for a brief period, dispenses order items into totes and then releases them for consolidation with other totes bound for the same destination.

In order to develop our simulation model in the most realistic manner, in particular, determining the model input values and the probabilistic distributions used to model the order picking process, we took advice from Kelton et al (2007), Banks (1998). Below are the two probability distributions we used and their main characteristics:

| Distribution | Parameters | Example Use |
|---|---|---|
| Exponential | Mean | Arrival times Time to machine failure |
| Triangular | Min, Mode, Max | Activity times |

The diagram below illustrates the order picking process within a "typical" crossdocking distribution centre.

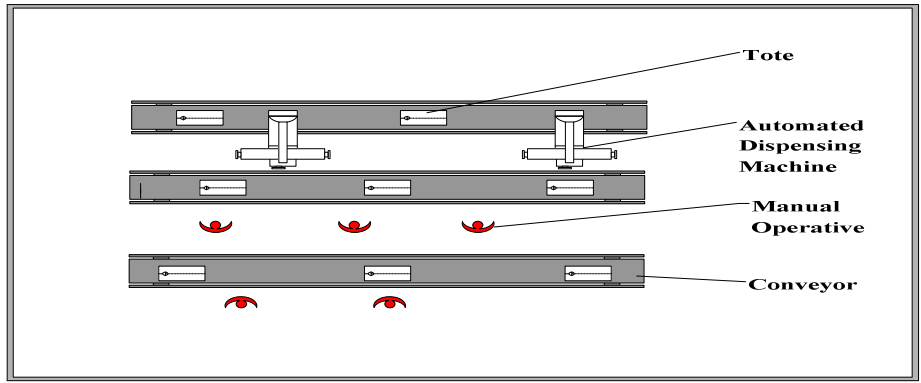

Figure 3. An order picking process within a typical crossdocking distribution centre.

### 3.4 SIMULATION MODEL VALIDATION

"Model validation is substantiating that the model, within its domain of applicability, behaves with satisfactory accuracy consistent with the study objectives" Balsi (1995) i.e. building a correct model for the intended purpose. For the purpose of validating our simulation model (Model 1), we used the Quantitative approach (Law and Kelton, 2000). as well as the Visual Check approach Robinson (1997).

Here is a brief description of the three experimental settings used for the validation process:

a. Model 1 (base model) entity arrival rate uses an exponential probability distribution and manual / automated order picking process use a triangular probability distribution.
b. Model 1-1 entity arrival rate uses an exponential probability distribution, manual / automated order picking process use a triangular probability distribution and it is enhanced by additional processing. This is station (buffer) processing which uses a triangular probability distribution.
c. Model 1-2 entity arrival rate uses an exponential probability distribution, manual / automated order picking process use a triangular probability distribution and is enhanced by additional processing. This is station (buffer) processing which uses a triangular probability distribution. This model, also implements the common random number technique (i.e. dedicating a different random number stream to sources of model variance Kelton et al (2007), (Law and Kelton, 2000).

We ran the model under three experimental settings, which involved varying the level of complexity of the simulation models, with a view to comparing the confidence interval of the differences in the means and the ratio of estimated variances. The model output data comparisons are in the following order, Model 1 and Model 1-1 and Model 1 and Model 1-2. The simulation output performance measure value of interest is "Total usage cost" and the fixed number of simulation replications is 500. We chose to run the simulation model over 500 replications because we did not know in advance the exact number of simulation replications that would give us a target confidence interval half width on a selected performance measure, which will be statistically robust for validation purposes Kelton et al (2007).

The Arena® simulation software has a built in statistical output analysis software, Output Analyser®, which was used for the validation experiments. We used its default, the Paired-t Test to compare the difference between each pair of mean observations across the two data files. As an alternative test, we could have used the Two-sample-t test, which is an approximation method that is used to generate the confidence interval on the difference in means. However, we choose to use the Paired-t Test because using the Two-sample-t Test will require the samples in the two data files to be statistically independent of each other; but for the default Paired-t Test, they could be correlated (e.g., from using common random numbers). See (Law and Kelton, 2000) for an explanation of these two tests on means and (Miller and Freund, 1977) for an explanation on the comparison of variances.

Below are the experiment results for the Quantitative validation approach:

- *Total Usage Cost* means and variance comparison results for *Model 1 - Model 1-1* (Quantitative validation approach, (Law and Kelton, 2000).

*Paired-T Means Comparison:*

| IDENTIFIER | ESTD. MEAN DIFFERENCE | STANDARD DEVIATION | 0.950 C.I. HALF-WIDTH | MINIMUM VALUE | MAXIMUM VALUE | NUMBER OF OBS |
|---|---|---|---|---|---|---|
| Total Usage Cost | 21.9 | 1.75e+003 | 154 | 1.49e+005 | 1.57e+005 | 500 |
| | | | | 1.49e+005 | 1.57e+005 | 500 |

FAIL TO REJECT H0 => MEANS ARE EQUAL AT 0.05 LEVEL

*Variances Comparison:*

| IDENTIFIER | VARIANCE RATIO | UPPER 0.950 C.I.LIMIT | LOWER 0.950 C.I.LIMIT | MINIMUM VALUE | MAXIMUM VALUE | NUMBER OF OBS |
|---|---|---|---|---|---|---|
| Total Usage Cost | 0.867 | 0.727% | 1.03% | 1.49e+005 | 1.57e+005 | 500 |
| | | | | 1.49e+005 | 1.57e+005 | 500 |

FAIL TO REJECT H0 => VARIANCES ARE EQUAL AT 0.05 LEVEL

Based on the results above, where the comparisons are in the direction of Model 1 to Model 1-1, we can see that the confidence interval of the differences in the means is 21.9. The ratio of estimated variances is 0.867 for the selected output performance measure (Total usage cost). The implication of this is that the difference is not significant at a mean maximum level of £157,000 and a mean minimum level of £149,000. The non significant difference can be explained by the changes in the models. The change in Model 1 and Model 1-1 is mainly, the introduction of the station (buffer) processing.

- *Total Usage Cost* means and variance comparison results for *Model 1 - Model 1-2* (Quantitative validation approach, (Law and Kelton, 2000).

*Paired-T Means Comparison:*

| IDENTIFIER | ESTD. MEAN DIFFERENCE | STANDARD DEVIATION | 0.950 C.I. HALF-WIDTH | MINIMUM VALUE | MAXIMUM VALUE | NUMBER OF OBS |
|---|---|---|---|---|---|---|
| Total Usage Cost | -14.6 | 1.87e+003 | 164 | 1.49e+005 | 1.57e+005 | 500 |
| | | | | 1.49e+005 | 1.57e+005 | 500 |

FAIL TO REJECT H0 => MEANS ARE EQUAL AT 0.05 LEVEL

*Variances Comparison:*

| IDENTIFIER | VARIANCE RATIO | UPPER 0.950 C.I.LIMIT | LOWER 0.950 C.I.LIMIT | MINIMUM VALUE | MAXIMUM VALUE | NUMBER OF OBS |
|---|---|---|---|---|---|---|
| Total Usage Cost | 0.974 | 0.817% | 1.16% | 1.49e+005 | 1.57e+005 | 500 |
| | | | | 1.49e+005 | 1.57e+005 | 500 |

FAIL TO REJECT H0 => VARIANCES ARE EQUAL AT 0.05 LEVEL

Based on the results above, where the comparisons are in the direction of Model 1 to Model 1-2, we can see that the confidence interval of the differences in the means is -14.6. The ratio of estimated variances is 0.974 for the selected output performance measure (Total usage cost). The implication of this is that the difference is not significant at a mean maximum level of £157,000 and a mean minimum level of £149,000. The non significant difference can be explained by changes in the models. The changes between Model 1 and Model 1-2 are mainly, the introduction of the station (buffer) processing and the implementation of the common random number technique (i.e. dedicating a different random number stream to sources of model variance Kelton et al (2007), Law and Kelton (2000)) within Model 1-2.

For the Visual Check validation, we conducted the experiment under the assumption that Model 1 will be run under "simpler" characteristics i.e. three types of entities instead of the default five types of entities, the replication length was set at 10 days instead of the default 30 days and the number of simulation replications was set at 100 instead of the default which is 500. Here is a summary of the steps performed for the Visual Check validation:

a. Stepping through the model to verify that different entities (types of orders) are being released into the system,
b. Stopping the model to verify that entities were following the correct processing sequence,
c. Separately enforcing variation in entity processing and reverse routing

conditions in a bid to see how the model will react.
d. Scheduling a fixed number of entities through the model to see that the exact number of entities is created, processed and disposed.

A visual display of the order picking simulation model while it is running is illustrated below:

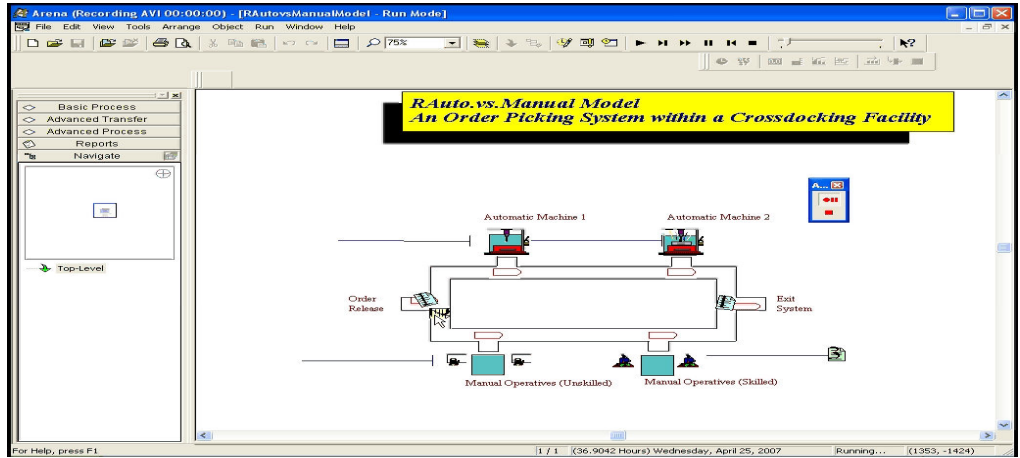

Figure 4. Simulation animation of an order picking process

As a result of our Visual Check validation test, we can conclude that the behaviour of the simulation model is reasonable, and closely resembles that which is expected for the purpose of our simulation model development.

**3.5 DESIGN OF EXPERIMENTS**

In simulation, design of experiments provides a way of structuring the environment within which experiments are performed. "It provides a way of deciding before the runs are made which particular configurations to simulate so that the desired information can be obtained with the least amount of simulation" (Law and Kelton, 2000). This can include what specific configuration we are interested in, the output performance of interest, right up to the way random numbers are being applied. For the purpose of our experiments, we made a single simulation run consisting of a fixed number of replications.

Here is a brief description of the three experimental settings used for the validation process:
   d. Model 1-S (base model) entity arrival rate uses an exponential probability distribution and manual / automated order picking process use a triangular probability distribution.
   e. Model 1-1-S entity arrival rate uses an exponential probability distribution, manual / automated order picking process use a triangular probability distribution and it is enhanced by additional processing. This is station (buffer) processing which uses a triangular probability distribution.
   f. Model 1-2-S entity arrival rate uses an exponential probability distribution, manual / automated order picking process use a triangular probability distribution and is enhanced by additional processing. This is station (buffer) processing which uses a triangular probability distribution. This model, also implements the common random number technique (i.e. dedicating a different random number stream to sources of model variance Kelton et al (2007), (Law and Kelton, 2000).

For our study, the experimental conditions are listed below:

a. *Simulation model resource configuration:*
1. Skilled Order Picking Operatives: Two at each order picking point
2. Unskilled Order Picking Operatives: Two at each order picking point
3. Automated Order Picking dispensers: One at each order picking point

b. *Number of order types (Entities):*

|  | Order type | Probability Distribution |
| --- | --- | --- |

|   |                                             | (%) |
|---|---------------------------------------------|-----|
| 1 | Many Items Few Quantities (MIFQ)            | 20  |
| 2 | Many Items Many Quantities (MIMQ)           | 25  |
| 3 | Few Items Few Quantities (FIFQ)             | 10  |
| 4 | Few Items Many Quantities (FIMQ)            | 15  |
| 5 | Regular Items Regular Quantities (RIRQ)     | 30  |

### 3.6 RESULTS

We used the same experimental settings as were used for the model validation tests for the Sequential Sampling method experiments. This is because just experimenting with Model 1 (base model) would not give us an extensive understanding of the behaviour of the order picking process i.e. source of variance, or provide a deeper insight into the performance of the Sequential Sampling method as a variance reduction technique for this class of Just in Time (JIT) systems. It is pertinent to mention that Model 1-S, Model 1-1-S, Model 1-2-S, respectively, are not different models but a single simulation model which has been extended in model complexity for experimental purpose. Our quantification of a reduction in variance by the application of the Sequential Sampling method is measured by a change in the confidence interval half width of the selected simulation output performance measure.

Based on the application of the Sequential Sampling method, these are the results we recorded for the Models 1-S, 1-1-S, 1-2-S, respectively. We achieved a 95% confidence interval half width on the selected simulation performance measure (total usage cost), of 2.80, 2.50 and 2.49, with average mean values are £115,000, £115,000, £117,000. This means in 95% of repeated trials, the average mean value for the selected simulation output performance measure value would be reported as within ± the half width. The number of simulation replications and quantity of simulation time over a single run required to achieve the above results for Models 1-S, 1-1-S, 1-2-S, respectively, are 568,802, 708376 and 717,922 and 30,240 minutes, 40,320 minutes and 43,200 minutes. There was a reduction on average over the three models in the 95% confidence interval half width for the selected simulation performance measure across the models of about 10.7%, however, it required on average about 20.2% more simulation replications, to achieve such an improvement. This improvement in solution quality i.e. reduction in the 95% confidence interval half width was due to an increased number of replications. We also discovered that by using common random numbers, there was no significant enhancement in the ability of the Sequential Sampling method to reduce variance within the selected simulation output performance measure for Model 1-1-S as compared with Model 1-2-S. Generally, Sequential Sampling method increases the precision of the selected simulation output performance measure but running Model 1-S at different levels of complexity using the Sequential Sampling method in order to reduce variance, requires a huge number of simulation replications. Here is a summary of our results:

Below are the assumptions upon which simulation model is based:
a. No Warm up period
b. Common Random Number Seed (Assigning a unique random number stream to sources of model variance i.e. Arrival process and Order picking process Kelton et al (2007), (Law and Kelton, 2000).
c. Simulation replication length: 16 Hours per day (8 Hour shifts), 30 Days (28800 minutes)
d. Proposed number of simulation replications: 999,999
e. Selected output performance measure: Total usage cost
f. Target output performance value: 95% Confidence interval ± half width of 0.5.

| Model Name   | Variance Measurement (Number of Replications) | Total Usage Cost £ | 95% C.I. Half Width | Real Simulation Time      |
|--------------|-----------------------------------------------|--------------------|---------------------|---------------------------|
| Model 1-S    | 568,802                                       | 115,000            | 2.80                | 30,240 minutes (21 days)  |
| Model 1-1-S  | 708,376                                       | 115,000            | 2.50                | 40,320 minutes (28 days)  |
| Model 1-2-S  | 717,922                                       | 117,000            | 2.49                | 43,200 minutes (30days)   |

Based on the summary of results above, a comparison of total usage cost between Model 1-S and Model 1-2-S, reveals a difference of 1.7%. The implication of this is that the difference is not significant at a mean maximum level of £117,000 and a mean minimum level of £115,000, when further compared with the gain in precision i.e. a lower 95% confidence interval half width of 2.49 as opposed to 2.80. The difference in mean values can be explained by a change in the models. The change in Model 1-S and Model 1-2-S is caused by the use of the station (buffer) processing.

## 4. CONCLUSIONS

We have investigated reducing variance associated with a simulation output performance measure (Total usage cost), using the Sequential Sampling method while applying minimum simulation replications, to achieve a desired 95% confidence interval half width of ± 0.5 for our chosen output performance measure. Our results indicate that, for our model, the Sequential Sampling method as a technique for variance reduction requires a lot of simulation replications to achieve the desired confidence interval half width and using common random numbers do not significantly enhance the performance of the Sequential Sampling method.

Future work to be undertaken is two fold. Firstly, we will be investigating the source of simulation model variance which has an influence on the simulation computational effort i.e. the number of replications required to achieve our target confidence interval half width. We will study the effect on output parameters of different combinations of input parameters and distributions, in particular model entity arrival rate, processing time, automated machine failure time and the different schedules of manual order pickers and automated order dispensing machines.

Secondly, we will also be developing a variance reduction technique which will enhance the performance of the Sequential Sampling method by adding a Particle Filter. This is an intelligent and complex model estimation technique based on simulation and is mainly used to draw samples from probability distributions. The Particle Filter is a widely used technique for estimating positioning and navigation Gustafsson et al (2002).

The main purpose for enhancing the performance of the Sequential Sampling method by adding a Particle Filter is to achieve a target confidence interval half width on a selected performance output measure using minimal number of simulation replications. Such an augmentation with the Particle Filter should enable the Sequential Sampling method to filter variance as the simulation replication is being performed and as such the simulation output performance value will have minimal variance and thus be more precise. This is the first time we are aware of a proposition to apply a Particle Filter to the Sequential Sampling method for the purpose of reducing variance to achieve a target confidence interval half width on a selected simulation output performance measure value.

**AUTHOR BIOGRAPIES**

**ADRIAN ADEWUNMI** is currently a PhD candidate of Computer Science and a member of the Automated Scheduling, Optimisation & Planning Research group (ASAP), School of Computer Science, University of Nottingham, Jubilee Campus, Wollaton Road, Nottingham, NG8 1BB, UK. http://www.cs.nott.ac.uk/~aqa

**PROF. UWE AICKELIN** is a Professor of Computer Science and a member of Automated Scheduling, Optimisation & Planning Research group (ASAP) School of Computer Science, University of Nottingham, Jubilee Campus, Wollaton Road, Nottingham, NG8 1BB, UK. http://www.cs.nott.ac.uk/~uxa

**DR. MIKE BYRNE** is a Lecturer in Operations Management at the Nottingham University Business School, University of Nottingham, Jubilee Campus, Wollaton Road, Nottingham,NG81BB,UK.http://www.nottingham.ac.uk/business/lizmdb.html